# Multi-Level Coding Efficiency with Improved Quality for Image Compression based on AMBTC

Dr.K.Somasundaram[1] and Ms.S.Vimala[2]

[1]Department of Computer Science and Applications Gandhigram Rural Institute
Gandhigram – 624 302
`somasundaramk@yahoo.co.in`
[2]Department of Computer Science Mother Teresa Women's University
Kodaikanal – 624 102
`vimalaharini@gmail.com`

## ABSTRACT

*In this paper, we have proposed an extended version of Absolute Moment Block Truncation Coding (AMBTC) to compress images. Generally the elements of a bitplane used in the variants of Block Truncation Coding (BTC) are of size 1 bit. But it has been extended to two bits in the proposed method. Number of statistical moments preserved to reconstruct the compressed has also been raised from 2 to 4. Hence, the quality of the reconstructed images has been improved significantly from 33.62 to 38.12 with the increase in bpp by 1. The increased bpp (3) is further reduced to 1.75in multiple levels: in one level, by dropping 4 elements of the bitplane in such a away that the pixel values of the dropped elements can easily be interpolated with out much of loss in the quality, in level two, eight elements are dropped and reconstructed later and in level three, the size of the statistical moments is reduced. The experiments were carried over standard images of varying intensities. In all the cases, the proposed method outperforms the existing AMBTC technique in terms of both PSNR and bpp.*

## 1. INTRODUCTION

Predictive Coding and Transform Coding are few image coding techniques that are widely used in the real time implementation of image and video coding [1]. Block Truncation Coding (BTC) is an efficient image coding method [2]. It finds its use in real-time image transmission due to its simplicity, performance and superior channel resisting capability [3]. Various steps are being taken by the researchers now-a-days in enhancing the BTC based techniques to improve them in terms of both PSNR and coding efficiency. In [4], Kumar and Singh presented a BTC called EBTC for higher PSNR that that of BTC and AMBTC. Wu presented a probability based BTC to reduce the bitplane overhead [5]. Wang and Chong proposed an adaptive multi-level BTC [6]. Somasundaram and Vimala developed an efficient block truncation coding by exploiting the feature of inter-pixel correlation [7]. Choi and Ko devised a novel DPCM-BTC [8]. Natarajan and Rao proposed two modified BTC algorithms by using the ratio of moments [9]. In BTC, input

DOI : 10.5121/ijist.2012.2204 35

image is divided into small blocks of size 4 x 4 pixels. For each block, the mean of the pixel values is computed using the equation (1).

$$\bar{x} = \frac{1}{m} \sum_{i=1}^{m} x_i \qquad (1)$$

where m is the total number of pixels in a block. The standard deviation $\sigma$ is computed using the equation (2).

$$\sigma = \sqrt{\frac{(\bar{x} - x_i)^2}{m}} \qquad (2)$$

Using the mean, the bitplane (B) of 0s and 1s is generated using the equation (3).

$$if\ x_i \geq \bar{x}\ then\ 1\ else\ 0 \qquad (3)$$

Two quantizers $q_1$ and $q_2$ are computed for each block using the equations (4) and (5).

$$q_1 = \bar{x} + \sigma\left(\sqrt{\frac{p}{m-p}}\right)\ for\ x_i \geq \bar{x} \qquad (4)$$

$$q_2 = \bar{x} - \sigma\left(\sqrt{\frac{m-p}{p}}\right)\ for\ x_i < \bar{x} \qquad (5)$$

The compressed image is stored or transmitted as a set of {B, $q_1$, $q_2$}. While reconstructing the image, the 1s in the bitplane are replaced with $q_1$ and 0s are replaced with $q_2$. The quality of the reconstructed image is computed with the Peak Signal to Noise Ratio (PSNR) using the equation (7). To compute the PSNR, the Mean Square Error (MSE) is computed using the equation (6).

$$MSE = \frac{1}{MxN} \sum (I_1(i,j) - I_2(i,j))^2 \qquad (6)$$

where MxN is the number of rows and columns of pixels in an image. $I_1$ is the input image and $I_2$ is the reconstructed image.

$$PSNR = 10xLog_{10}\left(\frac{255^2}{MSE}\right) \qquad (7)$$

Various BTC approaches have been proposed in the past [10]. Absolute Moment Block Truncation Coding (AMBTC) is an improved version of BTC, in which, instead of computing standard deviation ($\sigma$) that involves more multiplications, two quantizing levels high mean (hMean) and low mean (lMean) are computed using the equations (8) and (9).

$$hMean = \frac{1}{p} \sum_{i=1}^{p} x_i\ if\ x_i \geq \bar{x}. \qquad (8)$$

where p is the number of pixels in a block whose values are greater than or equal to mean.

$$lMean = \frac{1}{q} \sum_{i=1}^{p} x_i\ if\ x_i \leq \bar{x}. \qquad (9)$$

International Journal of Information Sciences and Techniques (IJIST) Vol.2, No.2, March 2012International Journal of Information Sciences and Techniques (IJIST) Vol.2, No.2, March 2012where q is the number of pixels in a block whose values are less than mean.

The PSNR obtained with the AMBTC is better than that of BTC. The improved variants of AMBTC: The Minimum Mean Square Error (MMSE). It is an iterative procedure to refine the quantizers that are generated [11] in consecutive iterations, the Minimum MAE Quantization (MMAE) [6] [12]. In another variant of BTC, called the New Look-up Table BTC Technique, the feature of inter pixel redundancy is used to reduce the bit rate further [13]. In EBTC [14], the blocks are categorized into high detail blocks and low detail blocks. For low detail blocks, only the mean alone is preserved thus leading to reduced bit rate. Like wise, much of study is conducted now-a-days to improve the BTC based techniques both in terms of bpp and the PSNR values.

In this paper, we have enhanced the existing AMBTC. In AMBTC, only two quantizers are used and hence the pixels in a reconstructed block take only either one of the two qantizers. In the proposed method (Improved AMBTC - IAMBTC), we have increased the number of quantizers to 4 and hence the number of bits used to represent the quantizing levels is raised. The remaining part of the paper is organized as follows: The proposed method is explained in Section 2, the results are discussed in section 3 and the conclusion is given in Section 4.

## 2. PROPOSED METHOD

In this method, the input image is divided into blocks of size 4 x 4 pixels. For each block, the high mean and the low mean, called the quantizers are computed using the equations (8) and (9) as in AMBTC. While encoding, to generate four quantizers, the step value is computed using the equation (10).

$$sv = (hMean - lMean)/3 \qquad (10)$$

The four quantizers, other than the hMean and lMean are computed using the equations (11) thru. (14).

$$Q1 = lMean \qquad (11)$$
$$Q2 = lMean + sv \qquad (12)$$
$$Q3 = Q2 + sv \qquad (13)$$
$$Q4 = hMean \qquad (14)$$

For each block, the bitplane is generated as follows: if the pixel value is closer to Q1, it is coded as 00. If the pixel value is closer to Q2, it is coded as 01, if closer to Q3, it is coded as 10 and if closer to Q4, it is coded as 11. Now a bitplane of size 32 bits is generated rather than 16 bits. As the quantizers Q2 and Q3 can be computed at the destination (decoding stage), only the two quantizers Q1 and Q4 are preserved along with the bitplane. The quantizers Q2 and Q3 are computed only when encding and decoding take place and not preserved along with the bitplane. The extended bitplane/quantizers are encoded in four different levels. In the first level, for each input image block, a bitplane of size 32 bits along with the quantizers Q1 and Q4 of size 16 bits are stored leading to a bit rate of 3 bpp.

As a step towards improving the coding efficiency, the feature of inter-pixel redundancy is exploited. Due to spatial redundancy, it is assumed that the nearest pixels will have more or less

37

the same intensity. Hence in the second level of compression, the 2$^{nd}$, 6$^{th}$, 10$^{th}$ and the 14$^{th}$ elements of the bitplane are dropped. The aforementioned elements are omitted as they can be regenerated by taking the average of the adjacent pixels. The above set {2, 6, 10, 14} of pixels can be dropped because they have pixels both in the left and right sides. The bpp obtained out of dropping the above mentioned elements of the bitplane is reduced to 2.5 bpp with only less acceptable degradation in the quality.

In the third level of reducing the bitrate further, the boldfaced elements as given in Fig. 1 are dropped. At the encoding stage, the dropped bits are regenerated using the equation set (15). As a result, the bpp is reduced to 2 with better PSNR when compared to AMBTC.

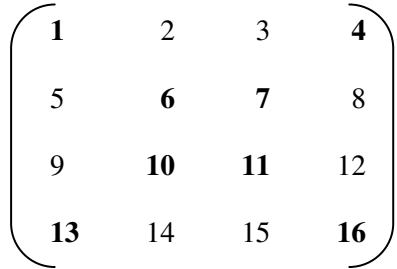

Fig. 1: Position of the elements to be dropped.

$$\left.\begin{array}{l} x_i = \dfrac{1}{2}(x_{i+1} + x_{i+4}) \quad \text{for i=1,11} \\ x_i = \dfrac{1}{2}(x_{i-1} + x_{i+4}) \quad \text{for i=4,10} \\ x_i = \dfrac{1}{2}(x_{i-1} + x_{i-4}) \quad \text{for i=6,16} \\ x_i = \dfrac{1}{2}(x_{i+1} + x_{i-4}) \quad \text{for i=7,13} \end{array}\right\} \quad (15)$$

In the fourth level of compression, the statistical moments Q1 and Q4 are divided by 4 to reduce the number of bits required to represent them. Generally it takes 8 bits to represent the gray level intensities of a gray scale image. As we divide the values by 4, the maximum value 256 can be transformed into 64 which require only 6 bits ($Log_2 64$ bits) to represent it. This leads to further reduction in bit rate leading to 1.75 bpp with onle negligible degradation in PSNR.

**Encoding Algorithm**

1: Input the image to be compressed.
2: Divide the image into small blocks of size 4 x 4 pixels.
3: For each input block, perform the following steps:
4: Compute the hMean and lMean using the equation (1)
5: Compute the four quantizer levels Q1, Q2, Q3 and Q4 using the equations (10) thru. (14).
6: Generate the Extended bitplane as
   b) if the individual pixel value is closer to
      i) Q1, code it as 00

  ii) Q2, code it as 01
  iii) Q3, code it as 10
  iv) Q4, code it as 11

6: Compression of the image is done in 4 different levels as follows:

Level1:   Extended bitplane of size 32 bits is generated and stored along with the quantizers Q1 and Q4.
                    or
Level2: Drop the elements in positions 2, 6, 10 and 14. Store or Transmit the reduced bitplane of size 24 bits along with the two quantizers Q1 and Q4.
                    or
Level3: Drop the elements as in Fig. 1. Store the reduced bitplane of size 16 bits along with the two quantizers Q1 and Q4.
                    Or
Level4: Level3 compression + divide the statistical moments by 4.

**Decoding Algorithm**

1: Input the Bitplane and the two quantizers: Q1 and Q4.
2: Compute the four quantizing levels using the equations through (10) thru. (14).
3:  Reconstruction of compressed image is done in three levels as follows:

Level1:  Reconstruct the image as decoding the elements of the bitplane as follows:
    a. if the element is 00, code it as Q1
    b. if 01, code it as Q2
    c. if 10, code it as Q3
    d. if 11, code it as Q4
                or
Level2: Recompute the dropped elements of the bitplane as follows:
    - 3 → average of 2 & 4
    - 7 → average of 3, 6 & 8
    - 11 → average of 10, 7 & 12
    - 15 → average of 14, 11 & 16
And perform Level2 actions
                or
Level3: Recompute the dropped elements using the equation set (15) and perform Level 2 actions.
                or
Level4: Compute Q1=Q1*4 and Q4=Q4*4 along with Level3 computations and perform Level2 actions.

## 4. RESULTS AND DISCUSSION

Experiments were carried out with standard images Cameraman, Boats, Bridge, Baboon and Lena of size 256 x 256 pixels. The input images taken for the study are given in Fig. 2.

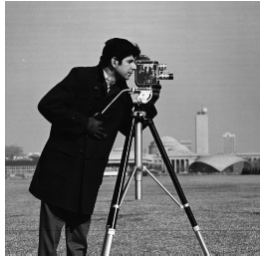 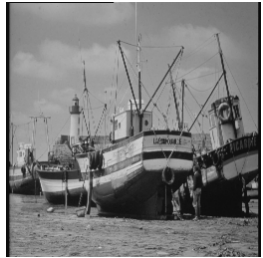 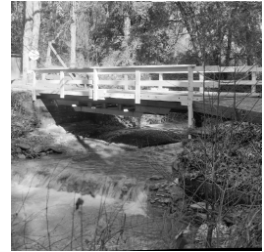

(a) Cameraman      (b) Boats      (c) Bridge

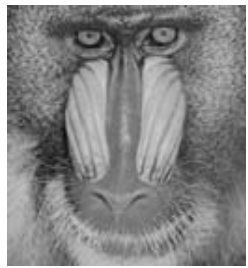 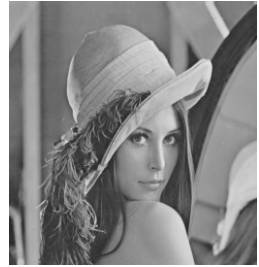

(d) Baboon      (e) Lena

Figure 2: Input images taken for the study

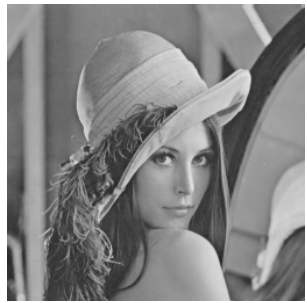 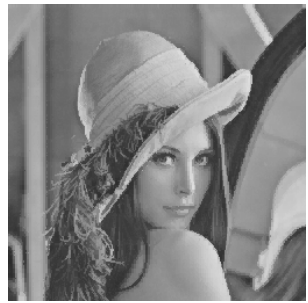

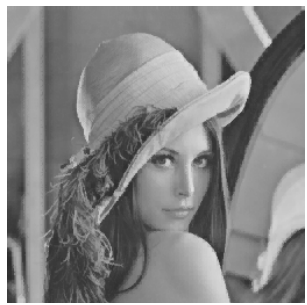

Original      AMBTC (34.90)      Level1 (38.82)
(bpp: 8)      (bpp: 2)      (bpp: 3)

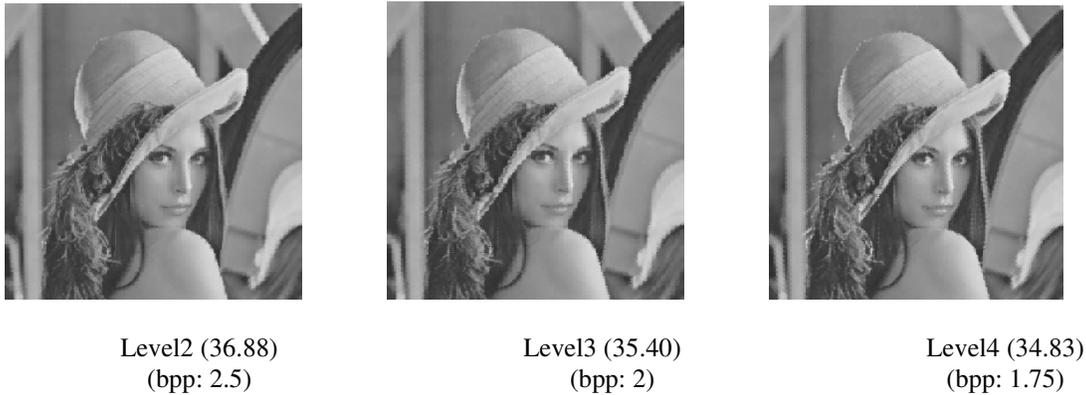

| Level2 (36.88) | Level3 (35.40) | Level4 (34.83) |
| (bpp: 2.5) | (bpp: 2) | (bpp: 1.75) |

Figure 2: Comparison of reconstructed Images with the AMBTC and the Proposed Methods

Table I presents the PSNR and the bpp obtained with the existing AMBTC Method and the proposed methods. The existing AMBTC method yields a bit-rate of 2 bpp. The average PSNR obtained is 33.62. With the proposed idea, the average PSNR obtained is 38.12 in Level1. The average PSNR has been raised by 4.50 on an average, which is significant improvement. But the bpp has been raised to 3 from 2. Hence further compression is done in three different levels. By adopting the interpolation method as in Level2, the bpp has been reduced from 3 to 2.5 and the average PSNR obtained is 36.88. Still to reduce the bpp, as in Level3, the elements as in Fig. 1 are dropped and regenerated. At this level, the average PSNR obtained is 35.40 and the bpp obtained is 2.

Further compression is achieved by adopting the Level4 technique. The average PSNR obtained is 34.83 with the bit rate of just 1.75. Finally the existing AMBTC when enhanced by incorporating the proposed idea in four different levels, gives better results both in terms of PSNR (34.83) and bpp (1.75).

The final results of the proposed method with bitplane compression and the moments compression are given in Table II.

The algorithms are implemented using Matlab 7.0 on Windows Operating System. The hardware used is the Intel Core 2 E7400@ Duo 2.8 GHz Processor with 2 GB RAM.

The reconstructed images using the AMBTC method and the proposed methods are given in Fig.3.

Table I: Comparison of PSNR and the bpp obtained with the proposed methods against the AMBTC method

| Image | AMBTC | | Level1 | | Level2 | | Level3 | | Level4 | |
|---|---|---|---|---|---|---|---|---|---|---|
| | bpp | PSNR | bpp | PSNR | bpp | PSNR | bpp | PSNR | bpp | PSNR |
| Cameraman | 2 | 32.17 | 3 | 38.60 | 2.5 | 37.22 | 2 | 35.48 | 1.75 | 35.28 |
| Bridge | 2 | 30.99 | 3 | 35.46 | 2.5 | 34.25 | 2 | 32.18 | 1.75 | 32.06 |
| Boats | 2 | 33.14 | 3 | 37.59 | 2.5 | 36.21 | 2 | 34.48 | 1.75 | 34.32 |
| Lena | 2 | 34.90 | 3 | 39.36 | 2.5 | 37.82 | 2 | 36.17 | 1.75 | 35.92 |
| Baboon | 2 | 36.90 | 3 | 39.61 | 2.5 | 38.90 | 2 | 36.96 | 1.75 | 36.59 |
| **Average** | **2** | **33.62** | **3** | **38.12** | **2.5** | **36.88** | **2** | **35.40** | **1.75** | **34.83** |

Table II: Improved PSNR obtained with the proposed method with the same bpp as that of AMBTC.

| Image | AMBTC | | Proposed Method (Level4) | |
|---|---|---|---|---|
| | PSNR | bpp | PSNR | 1.75 |
| Cameraman | 32.17 | 2 | 35.28 | 1.75 |
| Boats | 30.99 | 2 | 32.06 | 1.75 |
| Bridge | 33.14 | 2 | 34.32 | 1.75 |
| Lena | 34.90 | 2 | 35.92 | 1.75 |
| Baboon | 36.90 | 2 | 36.59 | 1.75 |
| **Average** | **33.62** | **2** | **34.83** | **1.75** |

## 4. CONCLUSION

The existing AMBTC method has been enhanced by increasing the number of quantizers from 2 to 4 and by increasing the size of the bitplane elements to 2. Ultimately, the bpp is increased. To have better coding efficiency, the bitplane is compressed in two different ways, and the size of the statistical moments is also decreased from 8 bits to 6 bits. Still there is a very good improvement in the PSNR values obtained with the proposed method. In existing AMBTC, the PSNR and the bpp obtained are 33.62 and 2 respectively. But with the proposed idea we can an average PSNR of 34.83 with just 1.75 bpp.

The proposed idea is incorporated with the existing AMBTC and the coding efficiency is improved in three different levels. In all the cases, we achieve better PSNR. Depending on the users' requirements, they can go for any one of the four different levels of compression.

# REFERENCES

[1]     Jain A.K., "*Image Data Compression: A Review*", Proc. IEEE, 1981, 69, pp. 349-380.

[2]      Delp E.J. and Mitchell O.R., "*Image Compression using Block Truncation Coding*", IEEE Transactions on Communication, Vol. COM-27, pp. 1335-1342, Sept. 1979.

[3]      Bing Zeng and Yrjo Neuvo, "*Interpolative BTC Image Coding with Vector Quantization*", IEEE Transactions on Communications, Vol. 41, No. 10, October 1993.

[4]     Kumar A. and Singh P., "Enhanced Block Truncation Coding for Gray Scale Image", International Journal of Comput. Tech. Appl., 2: 525-530, 2011.

[5]     Wu Y.G., "Block Truncation Image Bitplane Coding", Optical Engineering, 41: 2476-2478, 2002.

[6]      Wang J. ad Chong J.W., "Adaptive Multi-level Block Truncation Coding for Frame Memory Reduction in LCD Overdrive", IEEE Transactions on Consum. Elect. 56: 1130-1136, 2010.

[7]      Somasundaram K. and S.Vimala, "Efficient Block Truncation Coding for Image Compression", International Journal of Computer Science and Engineering, 2: 2163-2166, 2010.

[8]    Choi K.S. and KO S.J., "Improved differential Pulse Code Modulation – Block Truncation Coding Method adopting two-level Mean Squared Error Near Optimal Quantizers", Optical Engineering, 50: 47001-47007.

[9]     Natarajan S. and Rao Y.V.R., "Ratio Modified Block Truncation Coding Algorithm for Reduced Bitrates", Imaging Science Journal, 50: 25-31.

[10]    P.Franti, O.Navelainen and T.Kaukoranta, "*Compression of Digital Images by Block Truncation Coding: A Survey*", The Computer Journal, Vol. 37, No. 4, pp. 308- 332, 1994.

[11]    Madhu Shandilya and Rajesh Shandilya, "*Implementation of Absolute moment Block Truncation Coding Scheme Based on Mean Square Error Criterion*", SDR 03 Technical Conference and Product Exposition.

[12]    Lucas Hui, "*An Adaptive Block Truncation Coding Algorithm for Image Compression*", IEEE, PP.2233-2235, 1990.

[13]     Pasi Franti, Olli Nevalainen and Timo Kaukoranta, "*Compression of Digital Images by Block Truncation Coding:A Survey*", The Computer Journal, Vol.37, No.4, 1994.

[14]     K.Somasundaram and S.Vimala, "*Efficient Block Truncation Coding*", International Journal on Computer Science and Engineering Vol. 02, No. 06, 2163-2166, 2010.

## Authors

Dr.K.Somasundaram was born in the year 1953. He received the M.Sc. Degree in Physics from University of Madras , Chennai, India in 1976, the Post Graduate Diploma in Computer Methods from Madurai Kamaraj University, Madurai, India in 1989 and the Ph.D. Degree in Theoretical Physics from Indian Institute of Science, Bangalore, India in 1984. He is presently the Professor and Head of the Department of Computer Science and Applications, Gnadhigram Rural Institute, Gandhigram, India. From 1976 to 1989, he was a Professor with the Department of Physics at the same Institute. He was previously a Researcher at the International Center for Theoretical Physics, Trieste, Italy and a Development Fellow of Common Wealth Universities at the School of 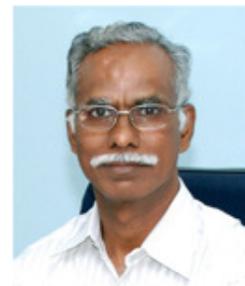 Multimedia, Edith Cowan University, PERT, Australia. His research Interests are in Image Processing, Image Compression and Medical Imaging. He has published more than 100 papers in National and International Journals like IEEE Transactions on Multimedia, Computers in Biology and Medicine (Elsevier), AEU, International Journal on Electronics and Communications, IEEE Transactions on Plasma Science, etc. He has attended several International conferences in USA, Australia, Italy, Singapore and Hungary. He has published three books: Programming in JAVA2 and Advanced Programming in JAVA2, (both by Jaico Publishing, Mumbai) are his best selling books. He is a Life Member of India Society for Technical Education and Telemedicine Society of India. He is also an Annual Member in ACM, USA and IEEE Computer Society, USA.

Ms.S.Vimala has received her M.Sc. degree in Computer Science at Bharathiar University, Coimabatore, India in 1994., M.Phil. Degree from Mother Teresa Women's University, Kodaikanal, India.. At present she is working as Assistant Professor in Department of Computer Science, Mother Teresa Women's University, Kodaikanal, India since 1999. Submitted her *Ph.D. thesis* on *Image Compression* in the month of March, 2011 and awaiting her results. She has published nearly 13 research papers in National/International Journals. She is a Life member of International Association of Computer Science and Information Technology (IACSIT). Her research interests are in Image Compression and Image Segmentation.

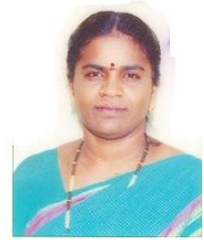